\documentclass{bmvc2k}

\title{Prior-Aware Synthetic Data to the Rescue:\\ Animal Pose Estimation with Very Limited Real Data}

\addauthor{Le Jiang}{}{1}
\addauthor{Shuangjun Liu}{}{1}
\addauthor{Xiangyu Bai}{}{1}
\addauthor{Sarah Ostadabbas}{https://web.northeastern.edu/ostadabbas}{1}

\addinstitution{
 Augmented Cognition Lab,\\
 Electrical and Computer Engineering Department,
 Northeastern University,\\ Boston, MA, USA
}

\runninghead{Jiang et al.,}{Prior-Aware Synthetic Data to the Rescue}

\usepackage{graphicx}
\usepackage{bbding}
\usepackage{tikz}
\usepackage{comment}
\usepackage{amsmath,amssymb} 
\usepackage{color}
\usepackage{amssymb}
\usepackage{pifont}
\newcommand{\xmark}{\ding{55}}%
\usepackage{booktabs}
\usepackage{multirow}
\usepackage{array}
\newcolumntype{Y}{>{\centering\arraybackslash}X}
\usepackage[accsupp]{axessibility}  
\newcommand{\eqnref}[1]{Equation~(\ref{eqn:#1})}
\newcommand{\figref}[1]{Fig.~\ref{fig:#1}}
\newcommand{\tabref}[1]{Table~\ref{tbl:#1}}

\newcommand{\overview}{
\begin{figure}
    \centering
    \includegraphics[width=1\linewidth]{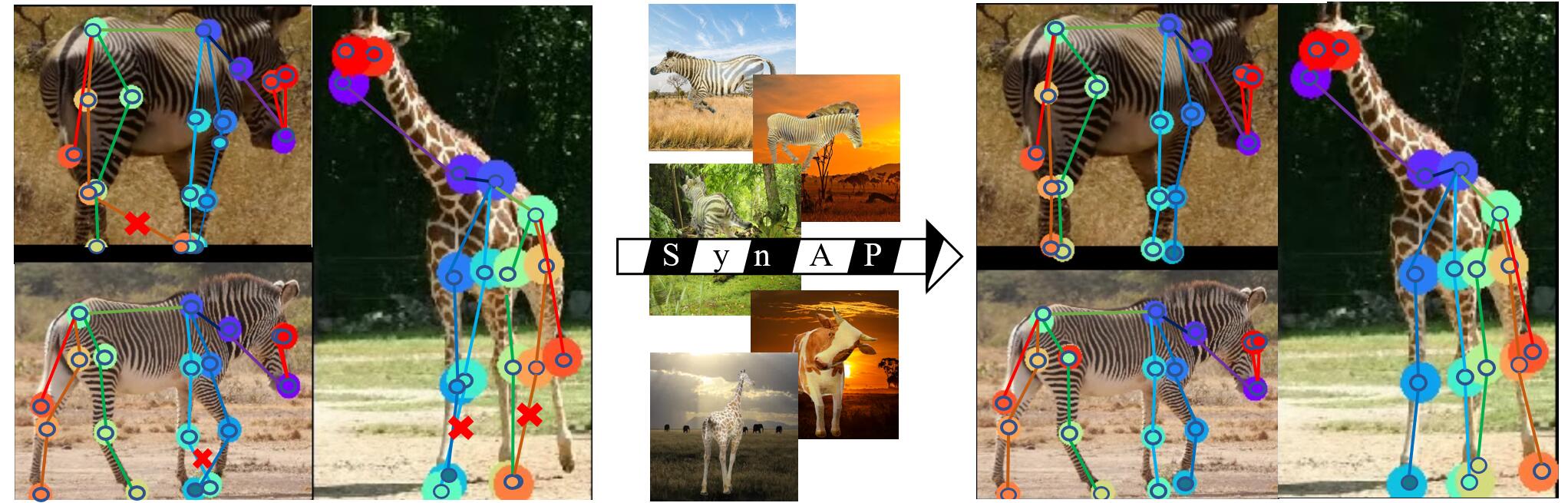}
    
    \caption{The outcome samples show the effect of model trained with or without synthetic animal pose  (SynAP) dataset. The left side is the pose estimation results based on the DeepLabCut \cite{Mathis_2021_WACV} pre-trained on ImageNet and fine-tuned by 99 real animal images and the right side shows the results of the same model when trained on the same amount of real images in addition to our SynAP dataset.  Wrong predictions are marked by red cross.} 
    \label{fig:overview}
    \vspace{-.1in}
\end{figure}
}

\newcommand{\deape}{
\begin{figure}[t]
    \centering
    \includegraphics[width=1\linewidth]{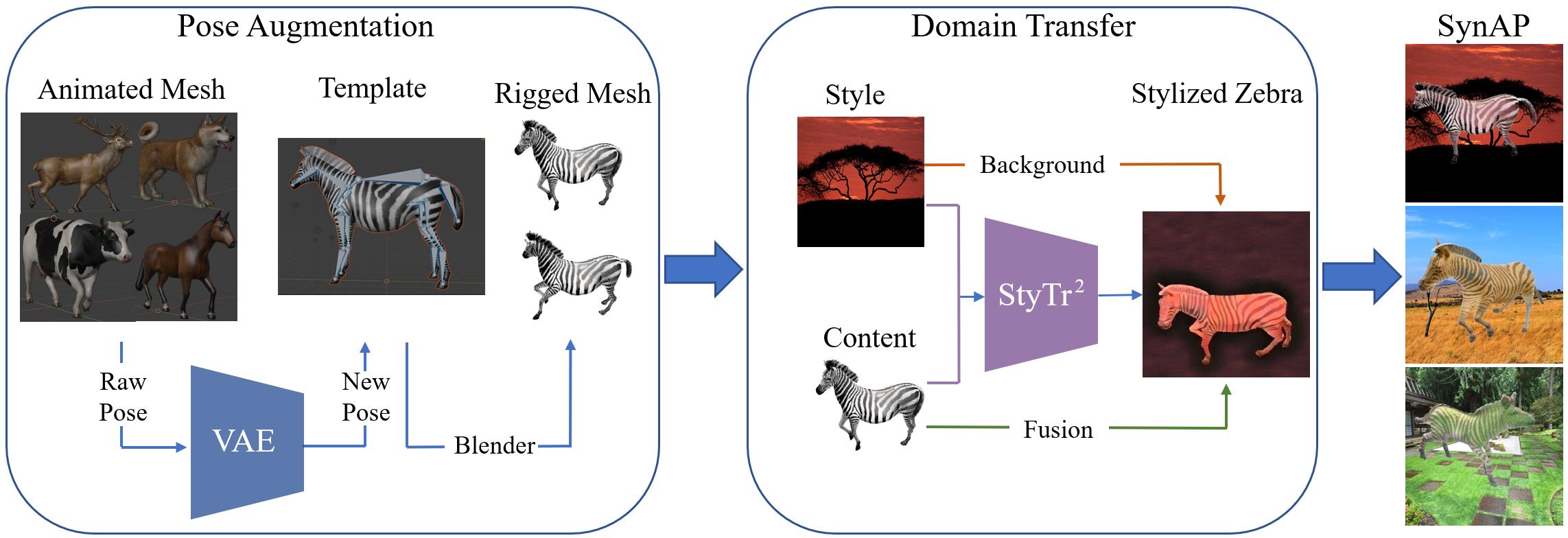}
    \caption{An overview architecture of our prior-aware synthetic data generation (PASyn) pipeline, composed of three parts: pose augmentation, domain transfer and dataset generation. The PASyn pipeline leads to generation of our probabilistically-valid synthetic animal pose (SynAP) dataset.}
    \label{fig:deape}
        \vspace{-.2in}
\end{figure}
}

\newcommand{\arrmature}{
\begin{figure}[t]
    \centering
    \includegraphics[width=0.8\linewidth]{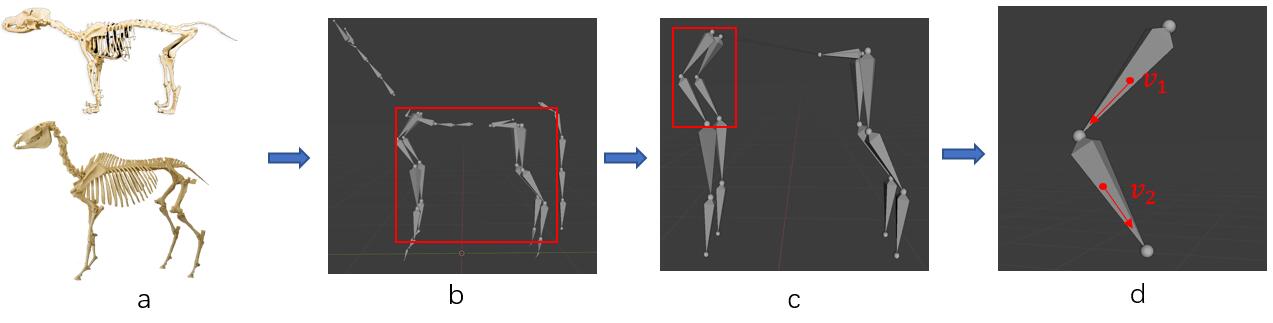}
    \caption{General quadruped armature. (a) shows the real skeletons of horse (bottom) and dog (top) respectively. (b) shows the artificially designed animal skeleton based on the real one. The red box marks the skeleton of limbs which we are interested in. (c) shows the skeleton of limbs. In (d), $v1$ and $v2$ represent the spatial orientation of two adjacent bones. The angle between the vectors is the angle between the two bones.}
    \label{fig:armature}
        \vspace{-.2in}
\end{figure}
}

\newcommand{\VAE}{
\begin{figure}[t]
    \centering
    \includegraphics[width=0.99\linewidth]{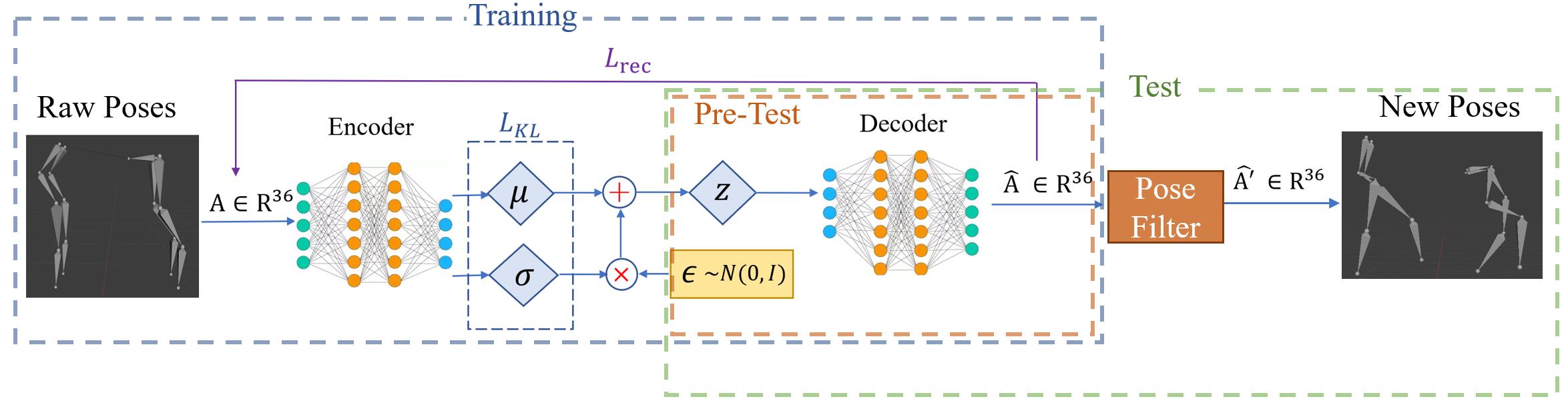}
    \caption{An overview architecture of our VAE-based animal pose generative model, composed of two main parts: training (blue-dot box) and test (green-dot box). Before we run ``test'' to generate new poses, pre-test (brown-dot box) should be performed independently to specify the sampling value ranges for each angles in the `pose filter'.}
    \label{fig:VAE}
            \vspace{-.2in}
\end{figure}
}

\newcommand{\posefilter}{
\begin{figure}[h]
    \centering
    \includegraphics[width=0.99\linewidth]{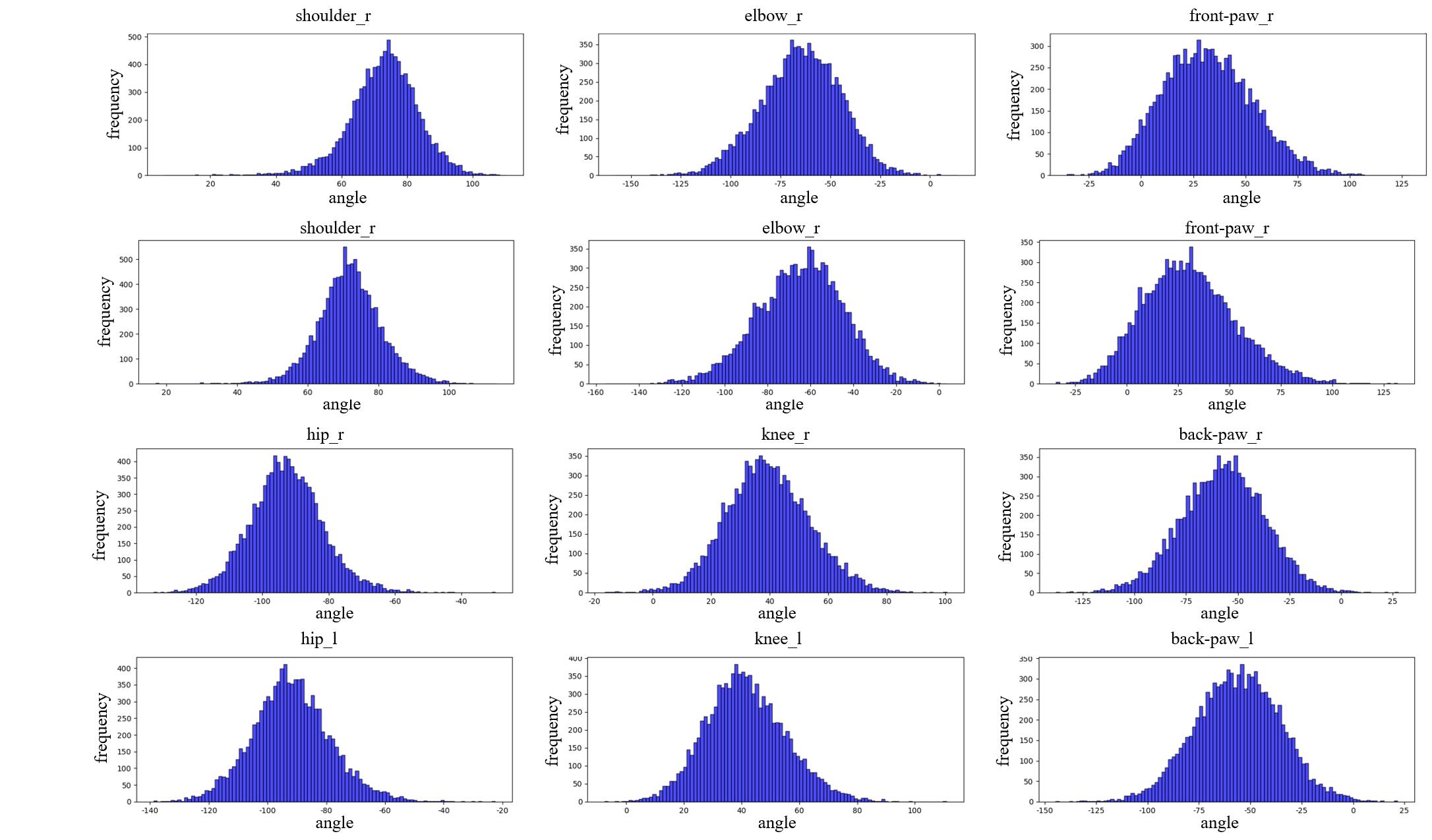}
    \caption{The histograms of angle values of 12 joints, which are shoulder\_right, elbow\_right, front-paw\_right, shoulder\_left, elbow\_left , front-paw\_left, hip\_right, knee\_right, back-paw\_right, hip\_left, knee\_left , back-paw\_left. }
    \label{fig:posefilter}
            \vspace{-.2in}
\end{figure}
}
\newcommand{\resultvisualization}{
\begin{figure}[h]
    \centering
    \includegraphics[width=0.70\linewidth]{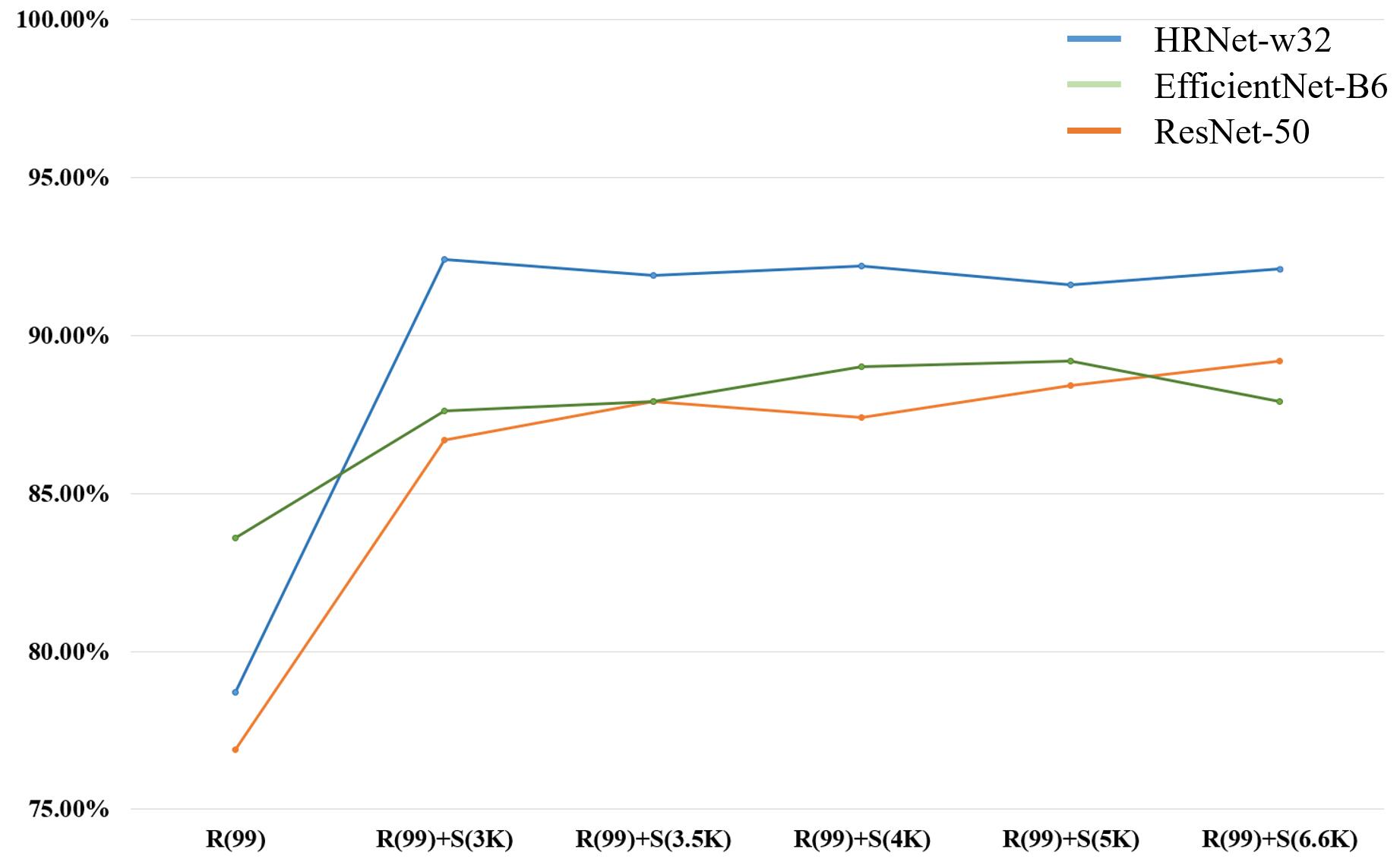}
    \caption{The line graphs showing the results of the three backbones, HRNet-w32 (in blue) \cite{hrnet}, EfficientNet-B6 (in green) \cite{tan2019efficientnet}, and ResNet-50 (in orange) \cite{resnet}, tested on Zebra-300 (300 real images) under different training sets. The training set contains only 99 real zebra images (from AP10K) and augmented with SynAP (3000 synthetic zebra images) or SynAP+ (3000  zebra and other animal synthetic images). We tried 4 different sizes of other animal synthetic images in SynAP+. There are 500, 1K, 2K, 3.6K.}
    \label{fig:resultvisualization}
            \vspace{-.2in}
\end{figure}
}

\newcommand{\zebracompare}{
\begin{table*}[t]
\caption{The effect of SynAP with limited real data of pose estimation results of three common backbones, HRNet-w32 \cite{hrnet}, EfficientNet-B6 \cite{tan2019efficientnet}, and ResNet-50 \cite{resnet} tested on Zebra-300 (300 real images). The training set contains only 99 real zebra images (from AP10K)  and augmented with SynAP (3000 synthetic zebra images) or SynAP+ (3000  zebra and 2000 other animal synthetic images). Best results for each backbone are shown in bold.}

\vspace{-.4cm}
\begin{center}
\resizebox{\textwidth}{!}{
\begin{tabular}{l l l |c c c c c c c c c c c}

\multirow{2}{*}{\textbf{Method}} &\multirow{2}{*}{\textbf{Backbone}} & \multirow{2}{*}{\textbf{Training Set}} & 
\multicolumn{11}{c}{\textbf{PCK@0.05 Pose Estimation Accuracy on Zebra-300 Set}}\\
\cline{4-14}
& & &Eye& Nose& Neck& Shoulders& Elbows&F-Paws&Hips& Knees&B-Paws&RoT&Average\\ \hline

\multirow{3}{*}{MMPose \cite{mmpose2020}}& \multicolumn{1}{|l}{ \multirow{3}{*}{HRNet-w32}}& \multicolumn{1}{|l|}{R(99)} & 97.3 &95.8 & \textbf{83.2}& 78.8&77.1 & 62.6&86.0&74.9&59.8 &82.4&78.7\\ \cline{3-14}
& \multicolumn{1}{|l}{}& \multicolumn{1}{|l|}{R(99)+ S(3K)}& \textbf{97.8}  &\textbf{98.3} &81.1 &\textbf{94.0} & \textbf{93.5}& \textbf{92.0}& 93.7&93.5&89.0 &\textbf{87.6}& \textbf{92.4}\\ \cline{3-14}
& \multicolumn{1}{|l}{}& \multicolumn{1}{|l|}{R(99)+ S(5K)}& 97.5  & 96.9&81.8 &89.6 &91.3 &90.7 &\textbf{94.1}&\textbf{94.1}&\textbf{90.4} &86.0& 91.6\\ \hline \hline

\multirow{3}{*}{DeepLabCut \cite{mathis2018deeplabcut}}&\multicolumn{1}{|l}{\multirow{3}{*}{EfficientNet-B6}} & \multicolumn{1}{|l|}{R(99)}& 93.7  & 96.2&\textbf{82.5} &\textbf{91.4} &80.8 &67.4 &88.1&84.5&71.8 &83.2&83.6\\ \cline{3-14}
& \multicolumn{1}{|l}{}& \multicolumn{1}{|l|}{R(99)+ S(3K)}& \textbf{95.1}  & \textbf{97.9}&81.5 &90.1 &83.3 &75.5 &\textbf{93.2}&89.3&83.9 &\textbf{86.8}& 87.6\\ \cline{3-14}
&\multicolumn{1}{|l}{} & \multicolumn{1}{|l|}{R(99)+ S(5K)}& 94.1  &92.6 &80.8 &90.8 &\textbf{87.0} &\textbf{85.7} &90.5&\textbf{93.3}&\textbf{88.3} &\textbf{86.8}&\textbf{89.2} \\ \hline
\hline
\multirow{3}{*}{MMPose \cite{mmpose2020}}&\multicolumn{1}{|l}{\multirow{3}{*}{ResNet-50}} &\multicolumn{1}{|l|}{R(99)}&96.2&96.9&\textbf{80.8}&59.0&71.3&71.2&88.5&78.2&59.3&85.2&76.9\\ \cline{3-14}
&\multicolumn{1}{|l}{} & \multicolumn{1}{|l|}{R(99)+ S(3K)}&95.6&95.8&69.9&\textbf{87.3}&\textbf{84.6}&84.3&90.8&91.2&84.4&77.2&86.7\\ \cline{3-14}
&\multicolumn{1}{|l}{} & \multicolumn{1}{|l|}{R(99)+ S(5K)}&\textbf{97.0}&\textbf{97.9}&74.5&85.3&84.2&\textbf{84.5}&\textbf{94.6}&\textbf{91.4}&\textbf{88.0}&\textbf{85.6}&\textbf{88.4} \\ \hline

\end{tabular}}
\vspace{-1cm}
\label{tbl:zebracompare}
\end{center}
\end{table*}
}

\newcommand{\result}{
\begin{table*}[h]
\caption{The effect of SynAP with large set of real data in pose estimation results of HRNet-w32 \cite{hrnet} tested on Zebra-300 (300 real images). The training set contains 8000 images from real animals (from AP10K) and augmented with SynAP (3000 synthetic zebra images) or SynAP+ (3000  zebra and 2000 other animal synthetic images). We tested the results when the 8000 images contain 99 real zebra inside them and when no zebra images is used. Please note that the HRNet-w32 model trained on AP10K is the state-of-the-art. Best results are shown in bold.}

\vspace{-.5cm}
\begin{center}
\resizebox{\textwidth}{!}{
\begin{tabular}{l c |c c c c c c c c c c c}
\multirow{2}{*}{\textbf{Training Set}} & \multirow{2}{*}{\textbf{Real Zebra}} & 
\multicolumn{11}{c}{\textbf{PCK@0.05 Pose Estimation Accuracy on Zebra-300 Set}}\\
\cline{3-13}
& &Eye& Nose& Neck& Shoulders& Elbows&F-Paws&Hips& Knees&B-Paws&RoT&Average\\ \hline
\multicolumn{1}{l|}{R(8K) (SOTA)}& \multirow{3}{*}{\checkmark }&\textbf{97.5}  &  97.2 & 79.4  & 87.8 & 90.3 & 93.8& 95.3 & 94.1 & 89.5 &86.4 &91.4\\ \cline{3-13}

\multicolumn{1}{l|}{R(8K) + S(3K)} & & 97.3& \textbf{98.3} & 79.0 & 93.1 & 94.9& 96.0 & 95.3&\textbf{96.7} &93.3&\textbf{89.6}&93.8\\ \cline{3-13}

\multicolumn{1}{l|}{R(8K) + S(5K)}&  &97.3 &97.6  & \textbf{81.1} & \textbf{93.7} &\textbf{95.7} & \textbf{96.0}&\textbf{96.6} &96.0 &\textbf{94.3} &87.6&\textbf{94.2}\\ \cline{3-13}  \hline
 \hline

\multicolumn{1}{l|}{R(8K)}& \multirow{3}{*}{\xmark}&79.7&87.7&37.4&77.6&80.0&87.6&82.0&86.4&81.3&67.2&78.3 \\ \cline{3-13}

\multicolumn{1}{l|}{R(8K) + S(3K)}&&94.8&96.2&\textbf{67.1}&90.8&87.9&90.2&87.6&91.6&89.7&77.6&88.2 \\ \cline{3-13}
\multicolumn{1}{l|}{R(8K) + S(5K)}&&\textbf{97.5}&\textbf{96.2}&66.1&\textbf{91.6}&\textbf{89.5}&\textbf{93.8}&\textbf{93.9}&\textbf{93.5}&\textbf{91.1}&\textbf{77.6}&\textbf{90.2}\\ \hline
\vspace{-.5cm}
\end{tabular}}
\label{tbl:result}
\end{center}
\end{table*}
}

\newcommand{\ablation}{
\begin{table*}[t]
\caption{Ablation study of two parts of our PASyn pipeline, VAE , style transfer and $\sigma^{2}$ random sampling distribution on two different zebra dataset with resolution of 300$\times$300. We use HRNet-w32 as our backbone here. The training set contains 99 real zebra (from AP10K) and augmented with SynAP (3000 synthetic zebra images). Best results are shown in bold.}

\vspace{-.3cm}
\begin{center}
\scalebox{0.7}{
\begin{tabular}{c|c|c|c|c|c|c}

\begin{tabular}[c]{@{}l@{}}\textbf{Index}\end{tabular}&
\begin{tabular}[c]{@{}l@{}}\textbf{Training Set}\end{tabular} & 
\begin{tabular}[c]{@{}l@{}}\textbf{VAE}\end{tabular} & 
\begin{tabular}[c]{@{}l@{}}\textbf{Style Transfer}\end{tabular} & 
\begin{tabular}[c]{@{}l@{}}\textbf{$\sigma^{2}$}\end{tabular} & 
\begin{tabular}[c]{@{}l@{}}\textbf{Zoo Zebra Set}\end{tabular} & 
\begin{tabular}[c]{@{}l@{}}\textbf{Zebra-300 Set} \end{tabular}  \\ \hline
a  & R(99) & \xmark & \xmark& 2I &76.0 & 78.7\\ \hline \hline
b  & S(3K) & \xmark & \xmark & 2I&38.7 & 30.0 \\ \hline
c  & S(3K) & \checkmark  & \xmark &2I &44.2& 36.7\\ \hline 
d  & S(3K) & \checkmark  & \checkmark  &2I&42.9 & 46.6\\ \hline \hline

e  & R(99)+S(3K) & \xmark & \xmark& 2I&89.8&88.0 \\ \hline
f  & R(99)+S(3K) & \checkmark  & \xmark &2I &90.4&89.8 \\ \hline
g  & R(99)+S(3K) & \checkmark  & \checkmark &I& 90.5& 91.1\\ \hline
h  & R(99)+S(3K) & \checkmark  & \checkmark  &2I& \textbf{91.5}&\textbf{92.4} \\ \hline

\end{tabular}
}
\vspace{-1cm}
\label{tbl:ablation}
\end{center}
\end{table*}
}
\newcommand{\supp}{
\clearpage
\newcommand{\beginsupplement}{%
        \setcounter{table}{0}
        \setcounter{equation}{0}
        \renewcommand{\theequation}{S\arabic{equation}}
        \setcounter{section}{0}
        \renewcommand{\thetable}{S\arabic{table}}%
        \setcounter{figure}{0}
        \renewcommand{\thefigure}{S\arabic{figure}}%
}

\newcommand{\independent}{\protect\mathpalette{\protect\independenT}{\perp}}
\def\independenT##1##2{\mathrel{\rlap{$##1##2$}\mkern2mu{##1##2}}}
\renewcommand\thesection{\Alph{section}}
\beginsupplement
\newcommand{\MYhref}[3][blue]{\href{##2}{\color{##1}{##3}}}

\section{Supplementary Materials}
\label{sec:supplementary}
In this section we offer figures and analyses supplementing our discussion from the main paper.

\subsection{Pose Filter}

Based on the \figref{posefilter}, which is made by decoding of 10,000 random samples from a normal Gaussian distribution, we can set a special sampling range for each joint. The angle value within this range can be regarded as a valid angle, and the pose that satisfies the twelve ranges can be seen as an appropriate pose. These ranges are shoulder\_right [40, 100], elbow\_right [-125, 0], front-paw\_right [-25, 100], shoulder\_left [40, 100], elbow\_left [-125, 0], front-paw\_left [-25, 100], hip\_right [-120, -60], knee\_right [0, 80], back-paw\_right [-125, 0], hip\_left [-120, -60], knee\_left [0, 80], back-paw\_left [-125, 0]. In order to increase the variety of poses and generate more poses with angles near the boundary, we choose to generate random samples from a Gaussian distribution $\mathcal{N}(0,2I)$ after setting the pose filter. The dropout rate of the pose is 68.0\%.

\posefilter

\newpage
\subsection{The Effect of Out-of-Domain Species on In-Domain Pose Estimation}
\resultvisualization
We explored the effect of the size of other synthetic animal data on target animal's pose predictions during training. The result is shown in \figref{resultvisualization}.

}

\begin{document}
\maketitle

\begin{abstract}
{
Accurately annotated image datasets are essential components for studying animal behaviors from their poses. Compared to the number of species we know and may exist, the existing labeled pose datasets cover only a small portion of them, while building comprehensive large-scale datasets is prohibitively expensive. Here, we present a very data efficient strategy targeted for pose estimation in quadrupeds that requires only a small amount of real images from the target animal. It is confirmed that fine-tuning a backbone network with  pretrained weights on generic image datasets such as ImageNet  can mitigate the high demand for target animal pose data and shorten the training time by learning the  the prior knowledge of object segmentation and keypoint estimation in advance. However, when faced with serious data scarcity (i.e.,  $<10^2$ real images), the model performance stays unsatisfactory, particularly for limbs with considerable flexibility and several comparable parts. We therefore introduce a prior-aware synthetic animal data generation pipeline called PASyn to augment the animal pose data essential for robust pose estimation. PASyn generates a probabilistically-valid synthetic pose dataset, SynAP, through training  a variational generative model on several animated 3D animal models. In addition, a style transfer strategy is utilized to blend the synthetic animal image into the real backgrounds.  We evaluate the improvement made by our approach with three popular backbone networks and test their pose estimation accuracy on  publicly available animal pose images as well as collected from real animals in a zoo\footnote{The SynAP dataset and PASyn code is available at \href{https://github.com/ostadabbas/Prior-aware-Synthetic-Data-Generation-PASyn-}{https://github.com/ostadabbas/Prior-aware-Synthetic-Data-Generation-PASyn-}.}.
}
\end{abstract}



\section{Introduction} 

 The research on animal pose estimation has grown in recent years, covering 2D/3D pose estimation, animal model recovery, and multi-animal pose estimation \cite{ap10k,mu2020learning,li2021synthetic,Mathis_2021_WACV,zuffi2019three}. The makeup of the available training sets for the animal pose estimation models divides the study into two branches. The first proposes training the model with significant amounts of labeled real data of a single species. 
The other one is based   on a small amount of real data of a target animal and more data from other adjacent domains to make up for the data scarcity of the target animal. Current data scarcity solution is centered around learning animals' common prior knowledge from large amounts of real data of similar species or even humans. Previous research has shown that quadrupeds, including humans, have similar appearances and skeletons, and that knowledge can be shared \cite{cao2019cross,Sanakoyeu_2020_CVPR,neverova2020continuous}.
This method is easier to implement after the largest labeled dataset for general animal pose estimation, AP10K \cite{ap10k}, was made publicly available in 2021. However, there are restrictions imposed by pose label divergence and the high cost of customization in the existing datasets. In contrast to human, there are huge varieties between animals with different length of bones, number of joints, and extra body parts. Large animal datasets with uniform labeling may leave out the special needs of researchers looking for creatures with unique physical traits.
\overview

Synthetic data is another potential choice of adjacent domain for the target animal when aiming for both "inexpensive" and "personalization" aspects. Once the synthetic animal model is built, the label can be defined on  the target model  and annotating pose data becomes considerably faster and less expensive. In addition to synthetic data, previous works \cite{mathis2018deeplabcut, ap10k} have proven that the backbone networks pre-trained on large general datasets such as ImageNet \cite{deng2009imagenet} will gain the prior of object segmentation and keypoint prediction. Thus, training even with few frames from the target leads to a high precision pose prediction. However, when there are far fewer images of the target available (e.g. less than 100), the size of the training set is insufficient for  a robust model fine-tuning. The accuracy of the model would decrease significantly when facing more actions from free-ranging behaviors, more self or environmental occlusions, and changes in the environment, textures, and shapes. \figref{overview} shows that how EfficientNet-B6 \cite{Mathis_2021_WACV} pre-trained on ImageNet and fine-tuned by 99 real animal images is error prone when tested on the images in the wild. 

To account for severe data scarcity  and guarantee a high degree of label personalization, we present a cost-effective and generic prior-aware synthetic data generation pipeline, called PASyn,  for animals pose estimate tasks that suffer from severe data scarcity. In short, this paper contributions are: (1) designing a novel variational autoencoder (VAE)-based synthetic animal data generation pipeline to generate probabilistically-valid pose data and verifying its performance on several animal pose test sets; (2) blending the synthetic animal images into  real backgrounds through style transfer to mitigate the inconsistency between synthetic and real domains; and (3) building a synthetic animal pose  (SynAP) dataset through PASyn, containing 3,000 zebra images and extending it with the 3,600 images of six common quadrupeds including, cows, dogs, sheep, deer, giraffes, and horses to make SynAP+ dataset; and (4)  releasing a new pose-labeled dataset of mountain zebras in zoo.

\section{Related Works}
It has been almost ten years since the early animal pose estimation work introduced in \cite{vicente2013balloon}. Yet, the performance of these models is still far inferior to the human pose estimation in terms of accuracy, cross-domain adaptation, and model robustness. This is mainly due to a lack of real-world data, which is the challenge for  almost all animal pose estimation work.

\subsection{Animal Pose Estimation with Label Scarcity} 
The most common statement in animal pose estimation articles is the lack of labeled dataset for training, which has been mentioned in many works \cite{deepgraphpose,cao2019cross,li2021synthetic,mu2020learning,mathis2018deeplabcut,graving2019deepposekit}. Numerable variety of species and subspecies, and considerable differences in physical characteristics and behavior patterns between them cause it difficult to form a labeled dataset with adequate samples. The cross-domain adaptation challenge exacerbates the situation. For each new animal, it is necessary to collect data from scratch and label them, since it is insufficient to learn the prior knowledge exclusively from the data of other animals. \cite{cao2019cross} proposed a cross-domain adaptive method and a large multi-species dataset, Animal-Pose \cite{Animalpose} to learn the shared feature space between human and animals, in order to  transfer the prior knowledge between them. \cite{ap10k} built their own large multi-species dataset, called AP10k \cite{ap10k} in order to train a robust and cross-domain model. However, both models are still unable to achieve the same level of 
pose prediction as animals in-domain when facing unseen species. 



\subsection{Animal Pose Estimation with Synthetic Data}
Synthetic data is a promising substitute for real data in previous works. Coarse prior can be learnt and then it would be used to build pseudo-labels for enormous amounts of unlabeled real animal data  \cite{mu2020learning,li2021synthetic}. The fly in the ointment is that works such as \cite{mu2020learning,li2021synthetic} still use significant amounts of real data (such as TigDog dataset \cite{TigDOg})  in training, which may not be possible to access for the unseen species. The work in \cite{zuffi20173d}, which focuses on animal model recovery, also gives an extraordinary hint on this problem. It purposed a general 3D animal model (called SMAL) by learning the animal toys' scan, and use the SMAL model and a few real pictures to generate a large amount of realistic synthetic data by adjusting the model's texture, shape, pose, background, camera, and other parameters. They also trained an end-to-end network \cite{zuffi2019three} using only synthetic Grevy’s zebra data, which came from direct reconstruction from animal 2D images. However, their results are much worse than the current state-of-the-art in terms of 2D animal pose estimation.

\subsection{Domain Gap between Real and Synthetic}
\label{domaingap}
Although synthetic data has advantages in terms of cost, it still faces the problem of fatal domain gap with the real data, which is even challenging to alleviate by current domain adaptation methods \cite{dann}. Pasting synthetic animal directly to the real background will result in the strong incongruity, mainly from the difference in projection, brightness, contrast, saturation, etc., between the two images. This incongruity can lead to excessive domain differences between synthetic and real data. Besides, to increase the texture diversity of synthetic animals to enhance the appearance robustness of the model, a common approach is to assign the synthetic data random textures from ImageNet. However, this further increases the discrepancy between synthetic and real domains and thus weakens the improvement of synthetic data in joint training of pose estimation tasks. The approach to increase model robustness and reduce domain variance by style transfer has been applied in medicine \cite{medical}, monocular depth estimation \cite{monodepth}, person re-identification \cite{reid}. We, therefore, adopt style transfer to alleviate the domain gap between the synthetic animal and the real background while increasing the texture diversity of the synthetic animal.
\section{PASyn Pipeline: Prior-Aware Synthetic Data Generation}

\deape
\label{sec:vdeap}
Our proposed prior-aware synthetic data generation (PASyn) pipeline enables robust animal pose estimation under severe label scarcity and divergence. The architecture of PASyn is shown in \figref{deape}, where we employ a graphic engine (i.e., Blender) to render the 3D animal mesh model into synthetic images. PASyn pipeline is composed of 3 key components: (1) pose data augmentation, (2) synthetic data stylization, and (3) synthetic animal pose image dataset generation. The details are presented in the following subsections. This paper presents PASyn with a working example of zebra as its target animal for pose estimation.


\subsection{Capturing Animal Pose Prior}
As evident based on the sample results shown in \figref{overview}, when the real target data provided to a pose estimation model is scarce, to improve the model performance, the prior knowledge of the target animal pose can be learned from a carefully curated synthetic dataset. To learn the animal pose priors, we use a variational autoencoder  (VAE) framework. VAE has already been proven to be useful for human pose priors learning \cite{vposer}, when it is trained on large-scale  3D human pose datasets. However, appropriate 3D datasets cannot be easily obtained for animal pose studies. Therefore, we feed the animation of multiple simulated animals made by human animators (inexpensively purchased from the CGTrader website), to the VAE, for it to learn the probabilistic distribution of the feasible poses of the targeted animals. Then, the trained VAE is used as a generative model to create our 3D synthetic animal pose (SynAP) dataset. As seen in \figref{armature}(a), the legs of quadrupeds like horses and dogs are similar in structure. This structure is obtained by simplifying the leg skeleton. Due to its applicability across quadrupeds, this structure is useful for training animal pose priors. Angles between neighboring bones (12 of them) in the structure shown in \figref{armature}(c) are chosen as data for training animal leg pose priors to minimise the influence of model sizes or bone lengths on training.

\arrmature

Our network follows the basic VAE process, including input, encoder, random sampling, decoder, and output as demonstrated in \figref{VAE}.  $A\in\mathbb{R}^{36}$ is a $ 1 \times 36 $ vector containing the angles between 12 pairs of adjacent bones, and each space angle is decomposed into three directions, and $\hat{A}$ is a vector of the same shape representing the generated angles as the model output. The role of the encoder which is composed of dense layers with rectified linear unit (ReLU)  activations is to calculate the mean $\mu$ and variance $\sigma$ of the network input. We use the reparameterization trick to randomly sample $z$ in latent space $Z\in\mathbb{R}^{16}$, where $z = \mu + \epsilon \times \sigma$ and $e$ satisfies the distribution in the shape of $\mathcal{N}(0,I)$. Finally, the decoder which has a mirror structure of encoder is used to reconstruct the set of space angle, and make the reconstruction results $\hat{A}$ as close to $A$ as possible. Considering that one infeasible angle can ruin an entire pose, during the pre-test, we first assign random sampling from a normal distribution to the variable $z$, and then we create the histograms of the 12 joint angles using enough $\hat{A}$ ($\approx$10,000 poses). Based on the statistical results, we can establish the sampling range for each angle. For each range, 5\% are excluded  of total angles, which are far from the mean value, and all of the ranges are recorded. After the pre-test, when we generate the new poses, a `pose filter' is designed to remove the pose $\hat{A}$ as long as there is an angle outside these specified ranges and finally produce the refined poses $\hat{A'}$.  We can also obtain poses with higher diversity by increasing the variance of the sampling distribution after determining the pose filter. The training loss of the VAE is:
\begin{align}
\label{eqn:kl}
    &\mathcal{L}_{\text {total}}=w_{1} \mathcal{L}_{KL}+w_{2} \nonumber \mathcal{L}_{\text {rec}},\\
    \mathcal{L}_{KL}=KL(q(z|A)||&  \mathcal{N}(0,I)), \,\,\, \text{and} \,\,\,  \mathcal{L}_{\text{rec}}=\|A-\hat{A}\|_{2}^{2},
\end{align}
where $w_{i}$ is the weight of each loss term. The Kullback-Leibler term, $\mathcal{L}_{KL}$, represent the divergence between the encoder’s distribution $q(z|A)$ and $\mathcal{N}(0,I))$.  $\mathcal{L}_{\text {rec}}$ is the reconstruction term. $\mathcal{L}_{KL}$ encourages a normal distribution noise while $\mathcal{L}_{\text{rec}}$, in contrast, encourages to reconstruct the $A$ without any divergence.

\VAE

\subsection{Stylization: Blending into the Background}
As shown in \figref{deape}, we use the large number of new synthetic poses obtained through VAE to rig the 3D zebra models and render synthetic animal images by changing parameters such as lighting and camera angles. Then, we assign a real (and context-related) background from forest scenes to each image. To alleviate the domain gap between the background image  $I_{sty}$ (style) and synthetic animal image $I_{cont}$ (content), while increasing the texture diversity of the synthetic animal we employ a style transfer technique. Unlike common convolution-only style transfer methods, we adopt an innovative image style transfer method, called StyTr$^2$ \cite{deng2022stytr2} based on multi-layer transformers. The $I_{sty}$ and $I_{cont}$ images are segmented into patches to generate feature embedding and eventually formed encoded content sequence by adding positional embedding for each patch. This process is calculated through the content-aware positional encoding based on the semantics of the image content, thereby eliminating the negative effect of scaling on the spatial relationship between patches. Two transformers are applied to encode the content sequence and style sequence, respectively. Then, a transformer decoder is utilized to decode the encoded content sequence according to the encoded style sequence in a regressive fashion.  Finally  a convolution-based decoder is used to decode the output sequence, consisting of a linear combination of encoded content and style sequence as well as positional embedding, to obtain the stylized content image $I^{sty}_{cont}$. Due to strong representation ability of transformers, this method can better capture accurate content representations and avoid loss of details than classical methods.

However,  StyTr$^2$ cannot freely adjust the intensity of style transfer, unlike the adaptive instance normalization (AdaIN) \cite{adain}. Besides StyTr$^2$, we also perform a simple pixel summation with normalization of the stylized synthetic animal data with the original data to control the intensity of style transfer with the fusion rate $\alpha$. The final stylized synthetic animal image is formulated as $I^{final}_{cont}  = (1- \alpha)I_{cont} + \alpha I^{sty}_{cont}$.

\subsection{Creating Synthetic Animal Pose (SynAP) Dataset}
\label{sec:datageneration}

Synthetic Animal Pose (SynAP) dataset contains 3,000 synthetic animal images generated by our prior-aware synthetic animal (PASyn) pipeline. Since zebra was selected as the main species for quantitative evaluation in this work, we collected 5 zebra toys and textured zebra 3D models and synthesized the images from these models with Blender.  3D synthetic model generation pipeline (3D-SMG) introduced in \cite{vyas2021efficient} was used to reconstruct two models from zebra toys, while the remaining 3 are adapted from 3D models. In addition to SynAP, which only contains zebra data, SynAP+ extends the SynAP with 3,600 images of 6 other animals, including horses, cows, sheep, dogs, giraffes and deer. We purchased one textured 3D model for each animal from a 3D model website, CGTrader and rendered 600 images for each of the aforementioned animals. The VAE model generates the animal pose in each image and 300 grass, savanna, and forest real scenes are collected from Internet to stylize the synthetic animal.The annotations will be automatically created with each image through calculating the coordinates of each zebra joint in the pose using bone vector transformations.

\section{Experimental Analysis}
We first describe implementation details of PASyn pipeline and different backbone models which we use to prove its effectiveness in animal pose estimation. In order to achieve a high prediction accuracy on an ``unseen'' animal in the existing labeled dataset with only a small amount of real data available, we select \textbf{zebra} as the main animal to verify the general effect of our PASyn pipeline on different models and quantify the result with the metric, PCK@0.05. we also perform an ablation study on each part of PASyn pipeline and finally show our model's generalization capacity on different species including deer, cows, dogs, sheep, horses, and giraffes. 

\subsection{Implementation Details}
\textbf{PASyn Pipeline:} We keep the setting in Vposer \cite{vposer} for training the VAE model. The learning rate of Adam optimizer is 0.001 and we set $w_{1}$ and $w_{2}$ which are the weights of $\mathcal{L}_{KL}$ and $\mathcal{L}_{\text{rec}}$ in \eqnref{kl}, as 0.005 and 0.01, respectively. The model is trained on 600 poses of animated realistic 3D models for 200 epochs with 128 poses in a batch. The models include horse, dog, sheep, cow, and deer. Each of these animals has over 20 common actions designed by 3D animators, such as running, walking, and jumping. To increase the diversity of the poses, we choose to generate random samples from a Gaussian distribution $\mathcal{N}(0,2I)$. The sampling value range of the pose filter will be described in the supplementary. For our domain transfer part, we use the pre-trained transformer decoder provided by \cite{deng2022stytr2} and set the fusion rate as 0.5.\\
\textbf{Backbone Pose Models:} We employ several state-of-the-art pose estimation structures with varying complexity as our backbone networks, including the ResNet-50 and HRNet-W32 of AP10K \cite{ap10k} and EfficientNet-B6 of DeepLabCut \cite{mathis2018deeplabcut} to reflect the general effect of our PASyn pipeline. All of the backbones are pre-trained on ImageNet \cite{deng2009imagenet}.


\subsection{Evaluation Datasets}

\textbf{Zebra-300 Dataset} is the primary test set we use. It contains 40 images from the AP10K test set, 160 unlabeled images randomly selected from AP10K, and 100 images from the Grevy's zebra dataset \cite{zuffi2019three}. We labeled the images of the latter two according to the AP10K dataset. The images in Zebra-300 dataset are mostly taken in the wild, so the environment occlusion makes the pose estimation task challenging. \\
\textbf{Zoo Zebra Dataset} is made up of pictures and videos we took at a zoo. The dataset contains 100 zebra images, including 2 different mountain zebra individuals. We performed manual animal pose labeling on the cropped and resized images following the label setting of the AP10K dataset. Each label has 17 key points: nose, eyes, neck, shoulders, elbows, front paws, hips, knees, back paws and root of tail. The occlusion of barbed wire makes this dataset more challenging. We will release this dataset along with the paper.


\zebracompare

\subsection{Pose Estimation Results}
The number of real (R) and synthetic (S) data shown in training set in \tabref{zebracompare} means the total number of images we used for model training. SynAP and SynAP+ are  divided into training and validation sets with a ratio of 7 to 1 as the AP10K setting. For real data, considering the data scarcity, we divide it with a ratio of 4 to 1. From \tabref{zebracompare}, we can clearly know that only using 99 images for training is insufficient for any backbone network. The highest prediction accuracy is 83.2\%, reached by  EfficientNet-B6. After adding SynAP to the training set, the prediction results dramatically increase to around 90\%. And after adding SynAP+ to the training set, the accuracy is further improved. Among them, HRNet-w32 achieves the highest accuracy of 92.4\% when trained with SynAP.

Different animals have domain similarity and the possibility of mutual transfer learning, which is proven in many works \cite{cao2019cross, ap10k}. Thus, we cannot ignore the existing labeled data, which can improve the prediction of unseen animal. In \tabref{zebracompare} and \tabref{result}, we can see that training the model with 99 zebra images and 8,000 images of the other animals can increase the average PCK from 78.7\% to 91.4\%. However, \tabref{zebracompare} shows that even with only 99 real we are still able to surpass that result and reach SOTA by adding SynAP and SynAP+. Moreover, \tabref{result} points that the models trained with 8,000 real and SynAP or SynAP+ is further improved to 93.8\% and 94.2\%. When there is no target animal (i.e., zebra) in the training set, while the state-of-the-art model suffers from this situation (accuracy drops to 78.3\%), the model can recover its performance if given the SynAP dataset, as \tabref{result} shows. This confirms that our method can achieve high-precision prediction even for the unseen animals.

\result

\ablation
\subsection{Ablation Study}
As seen from \tabref{zebracompare}, even if only trained on a small amount of real data, the model can accurately predict the keypoints with apparent features and less flexibility, such as nose and eyes. Various alternative candidate positions and large degrees of freedom of keypoints on the limbs would considerably affect the prediction results. The main improvement brought by SynAP and SynAP+ is the significant increase of the prediction of keypoints in animal limbs. To verify that the VAE, StyTr$^2$ models \cite{deng2022stytr2}, and the higher variance of VAE sampling distribution in our PASyn pipeline can mitigate the mismatching prediction of the limbs and reduce the domain discrepancy between the real (R) and synthetic (S) domains, we test them on the Zoo Zebra and Zebra-300 datasets. \tabref{ablation} shows the outcomes of seven experiments (a-g) with various training conditions. When the model trained without VAE, we will manually set a reasonable value range for the angles between adjacent bones on the limbs, and conduct randomly sampling, similar to the method used in \cite{zuffi2019three,mu2020learning}. We will keep the original texture of the model when  StyTr$^2$ is not used.
The various training and test sets clearly reflect that the model trained with both of VAE and StyTr$^2$ in our PASyn pipeline can achieve obviously higher accuracy than the model trained without them. Also, the effect of variance of the random sampling distribution, $\sigma^{2}$ on the pose estimation performance is shown in experiments (g) and (h). The higher diversity of the poses can improve the pose estimation, hence our choice of $\sigma^{2}= 2I$. 

\section{Conclusion}
 We presented a cost-effective and general prior-aware synthetic data generation pipeline (PASyn) for animal pose estimation, for target animal studies that suffer from a severe data scarcity. A probabilistic variational generative model as well as a domain adaptation technique are introduced to increase the validity of the generated poses  and to reduce the domain discrepancy between the synthetic and real images. Our synthetic animal pose dataset SynAP and its extended version SynAP+, and  the positive effect of them on pose estimation task of animals with a small amount of real data is verified on different backbones and achieves state-of-the-art performance.

\bibliography{ref}
\newpage
\supp
\end{document}